# Brain Tumor Segmentation Based on Refined Fully Convolutional Neural Networks with A Hierarchical Dice Loss

Jiachi Zhang, Xiaolei Shen, Tianqi Zhuo, Hong Zhou∗
Key Laboratory for Biomedical Engineering of Ministry of Education,
Zhejiang University of China

December 25, 2017

**Abstract**

As a basic task in computer vision, semantic segmentation can provide fundamental information for object detection and instance segmentation to help the artificial intelligence better understand real world. Since the proposal of fully convolutional neural network (FCNN), it has been widely used in semantic segmentation because of its high accuracy of pixel-wise classification as well as high precision of localization. In this paper, we apply several famous FCNN to brain tumor segmentation, making comparisons and adjusting network architectures to achieve better performance measured by metrics such as precision, recall, mean of intersection of union (mIoU) and dice score coefficient (DSC). The adjustments to the classic FCNN include adding more connections between convolutional layers, enlarging decoders after up sample layers and changing the way shallower layers' information is reused. Besides the structure modification, we also propose a new classifier with a hierarchical dice loss. Inspired by the containing relationship between classes, the loss function converts multiple classification to multiple binary classification in order to counteract the negative effect caused by imbalance data set. Massive experiments have been done on the training set and testing set in order to assess our refined fully convolutional neural networks and new types of loss function. Competitive figures prove they are more effective than their predecessors.

## 1 Introduction

Less common but indeed fatal, brain tumor is one of the most notorious medical threat to human beings [1]. Patients with the most aggressive tumors have a life expectancy of less than two years [2]. Brain tumors can be classified to be either primary tumor or metastatic tumors by its origin. Gliomas, as one of the most frequent primary tumors are the main object studied in the field of brain tumor segmentation. On the other hand, gliomas are graded into 4 levels by the World Health Organization (WTO). Grades I-IV denote the escalation of gliomas' aggressiveness. Commonly we call gliomas of grade I and II Low Grade Gliomas (LGG). Gliomas of grad III and IV are named as High Grade Gliomas which comprise of malignant gliomas and mostly cause the death [1], [3]. The most widely used therapies include surgery, radiotherapy and chemotherapy.

In order to provide more comprehensive information for diagnosis, the technology of brain imaging has been highly developed. Nowadays, prevalent methods such as Computed Tomography (CT), Magnetic Resonance Spectroscopy (MRS) and Magnetic Resonance Imaging (MRI) reconstruct human brain in 3 dimensional space and can well present different tissue information. MRI is a non-invasive technique which will cause little radioactive damage to brain tissue [1], [4], [5]. Its wide application ensures it can be voluminously collected thus satisfy the huge training demand from computer vision. Another reason MRI gains preference from researcher of medical image process is that it can provide superior contrast information between different brain tissues. Furthermore, different types of high contrast MRI image including T1-weighted MRI (T1), T2-weighted MRI (T2), T1-weighted MRI with gadolinium contrast enhancement (T1c) and Fluid Attenuated Inversion Recovery (FLAIR) all have their own advantage and disadvantage to delineate certain subregions of brain soft tissue and are usually combined to give a integral and hierarchical description for diagnosis [1], [3], [6], [7]. Figure 1 shows these 4 types of MRI image from one patient. However, dislike traditional semantic segmentation which can be easily finished by ordinary people, brain tumor segmentation is complex, tedious and acquires professional medical knowledge. Gliomas and some of its subregions exhibit an ambiguous boundary which is difficult to specify [7]. On the other hand, tumors' shape, location and modality are highly variable form patient to patient and the images' sequential qualities are uncertain along every axis of the 3D MRI image. Even experts with clinical experience for decades need to be discreet when they analyse MRI images. Researchers from computer vision have been taking effort to develop automatic or semi-automatic segmentation methods as auxiliary tools for brain tumor diagnosis. Since the proposal of deep learning, researchers are pursuing the goal of developing an artificial intelligence which could outperform physicians in this time consuming task of brain tumor segmentation.

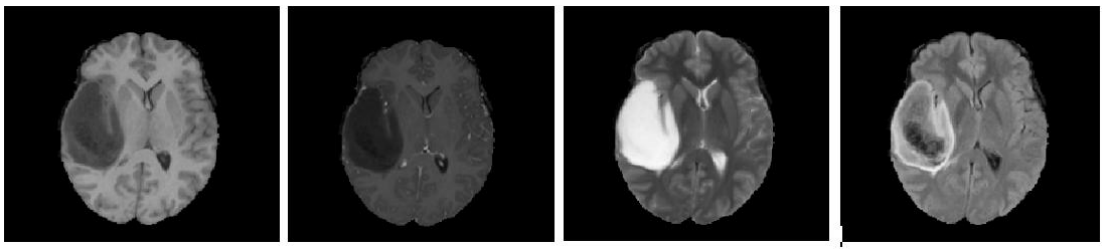

**Figure 1:** Serial slices from a patient. From left to right: T1-weighted MRI (T1), T1-weighted MRI with gadolinium contrast enhancement (T1c), T2-weighted MRI (T2) and Fluid Attenuated Inversion Recovery (FLAIR).

Traditional image process algorithms and machine learning algorithms have been applied to brain tumor segmentation since long. Havaei et al. [9] proposed a semi-automatic classification method using support vector machine (SVM). D Kwon et al. [10] proposed a generative model which generate tumors and edemas from several seed points with a priority to tumor shape. They achieved top raking in BRATS 2013 challenge. D Zikic et al. [11] differentiate brain tumor and its subregions with a discriminative approach based on decision forests. However, convolutional neural networks (CNN), as a powerful tool to extract features from raw images automatically, start to draw world wide attention and be soon applied to medical fields [2], [6], [12], [13], [14].

Inspired by preceding image classification CNN (AlexNet, VGG net) [15], [16] , patch-wise prediction can be transferred to be a good segmentation method as the patch will be centered to one pixel. Sérgio Pereira et al. [14] demonstrated this method obtained second position with DCS metric in BRATS 2015 challenge. However, this method is time consuming and less efficient because of its cropping pre-processing and a huge amount of data to process. Like the flatten vector can present a single image and can be classified by logistic or softmax classifier, researchers suggested that every pixel in a feature map retains whole information of its receptive field thus throughout up sample layers we can obtain a pixel-wise feature map without any fully connected layers [17]. In the recent 2 years, variants from the basic FCNN have been proposed. SegNet [18], an encoder-decoder architecture, retains the max-polling indices to better restore full resolution feature map. U-Net [19], which was designed for medical image, concatenates downsample layers and its counterpart of up sample layers. These networks mentioned above emphasize interconnections between downsample layers and up sample layers. Other works also show more connections between adjacent layers elevate networks performance, deepen the architecture and make the model easy to train [20]. In this paper, we try to add more connections to the basic FCNN both in the downsample path and in the up sample path and we prove the model gains higher accuracy in every time mark of training iteration.

Another adjustment is parallel structure or cascaded structure. Researchers have realized that on the one hand, we need to add max-pooling layer to enlarge the receptive field and add non-linearity to increase the accuracy, but on the other hand max-pooling layer decreases localization precision because it destroys images' spacial structure. Therefore, new types of FCNN with a path of multiple streams has been proposed [21]. One of the stream retains the localization information and another stream includes max-pooling layers. Inception [22], namely GoogLeNet, which was previously proposed, tried to solve the conflict by using different size of convolutional kernels and concatenating them together. Similarly, M Havaei et al. [2] proposed a Two-pathway CNN architecture which combines global and local information in the output layer. Besides the parallel structure, they also gave 3 cascaded architectures which differentiate each other by the concatenating locations. G Wang et al. [7] proposed a different types of cascaded network which made use of the brain tumors' hierarchy segmentation quality. Their experiments show parallel and cascaded structure work well on brain tumor segmentation.

Despite of the procedure of feature extraction, it is also important to pick up an effective classifier and especially a loss function to balance the unpaired class frequencies in the sample space. In the data set of BRATS 2015, taking the LGG set as an example, the ratio of tumor and non-tumor is closed to 87:1. Even the 4 types of brain tumor tissues which were officially labeled have a ratio of approximately 2:16:7:1. So the model is hard to converge if we do not take any measures. As variants from sigmoid cross-entropy loss for a binary classification, new types of loss may derive from evaluation metrics were introduced by Carole H Sudre et al. They also made comparisons between these loss functions and gave their recommendations. Lucas Fidon et al. gave an presentation of a multi-class dice loss using Wasserstein distance and applied it to a holistic convolutional network with a multi-scale fused loss function. In this paper, we propose a new type of classifier which transfers multiple classification to multiple binary classification using proper loss function and threshold value. We also use down sample method to manually balance the data sets. Specific approach and experiments will be explained in section 3 and 4.

## 2 Data Set

The data sets fed to training and testing procedure were collected from BRATS 2013 and BRATS 2015. It can be divided into two parts. High Grade Gliomas (HGG) data set includes 220 patients' MRI images and Low Grade Gliomas data set includes 54 patients' MRI images. Every patient has a group consisting 4 types of MRI brain images, T1, T2, T1c and FLAIR. Different from traditional RGB images, the MRI images have another axis thus have a shape of 155x240x240 but only have one channel. As the previous works indicated, the four types of MRI images should be combined into a 4 channel image. Therefore, the final shape of a single training material is 155x240x240x4.

In the training data set, a group of MRI images also includes a label image in which every pixel is given a label denoted by integer from 0 to 4. Figure 2 shows how different subregions of a brain tumor combine with each other and what each part represents. However, the target segmentation task in the challenge is not separating these 5 classes of brain tissue but classifying each pixel into classes overlapping or evolving with multiple classes. The aim can be described as following. First, detect the complete brain tumor which consists of necrosis, edema, non-enhancing tumor and enhancing tumor. Second, find the tumor core which comprises of necrosis, non-enhancing tumor and enhancing tumor. Finally, point out the enhancing tumor among from tumor core. Their relationship is demonstrated in Figure 3.

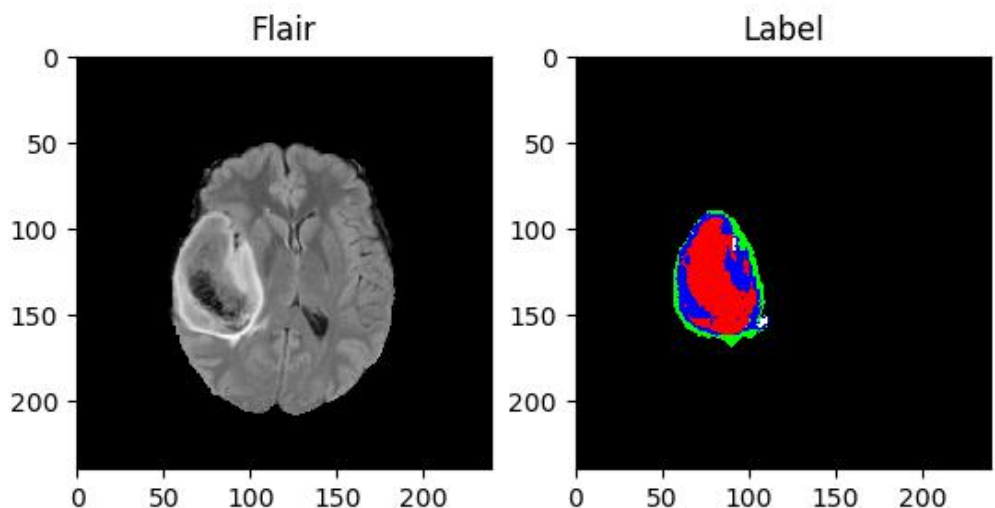

**Figure 2:** Left image shows a slice of a FLAIR MRI image. Right image shows the labels paired to the left. Red region (class 1): necrosis, green region (class 2): edema, blue region (class 3): non-enhancing tumor, white region (class 4): enhancing tumor.

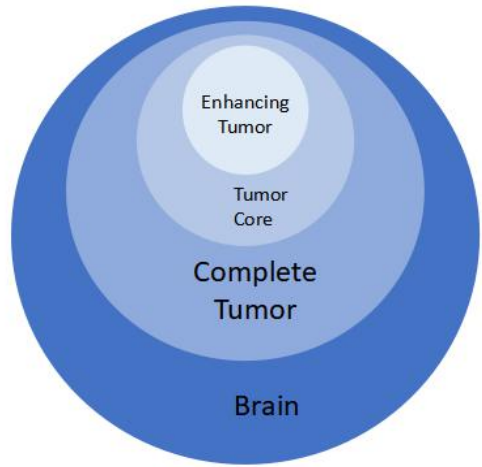

**Figure 3:** The relationship between 4 target classes. Although the 5 labeled classes are independent with each other. The 4 target classes share a containing relationship with each other. Tumor core contains enhancing tumor, non-enhancing tumor and necrosis. Complete tumor contains tumor core and edema.

## 3 Method

We divide this section into 2 parts. The first part discusses the procedure of feature extraction where we concentrate on the FCNNs' architecture. We apply several classic CNN architecture to our 2D slices of the 3D brain images. We try to achieve an end-to-end prediction so we abandon the fully connected layers. We describe the architecture as an encoder-decoder structure which is from a previous paper. The encoder denotes the down sample path and the decoder represents the counterpart up sample path to restore the full resolution. We make adjustments towards the encoder and the decoder respectively. In the second part, we apply new types of loss functions including bootstrapping loss, Dice Loss (DL) and Sensitivity-Specificity Loss (SSL). We make the comparisons between these loss functions and traditional cross-entropy loss. And we propose a hierarchical dice loss function inspired by the containing relationship of the 4 target classification.

### 3.1 FCNN Architecture

VGG16 + skip connection structure As a classic structure which won the ImageNet challenge in 2014, VGG was soon applied to a fully convolutional neural network [17]. Without upper three fully connected layers, the finally feature map was achieved by appending up sample layers and convolutional layers with 1x1 kernels. The author in that paper also adopted a skip connection structure to add up feature maps from different up sample layers. Figure 4 shows the skip connection structure. In practice, we adopt the so-called FCN-8s. The up sampled feature map is resulted from a transposed convolutional layer whose kernel is initialized by bilinear interpolation and should be learned to give finer segmentation quality from the coarse down sampled feature map. The initialized kernel can be described by the formula below:

$$F(x,y) = (1-\left|x-\frac{K-1}{2}\right|) \times (1-\left|y-\frac{K-1}{2}\right|)$$

where x, y are the filters' indices and K is filter size. x, y, K are unsigned integers.

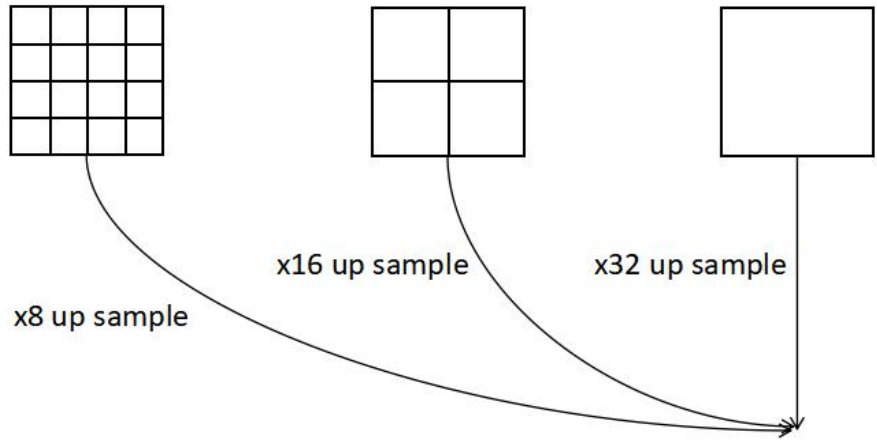

**Figure 4:** Skip connection structure as decoder. The 3 rectangles from right to left represent last 3 max-pooling layers of VGG16. After the up sample layers, they are added up with weights of 1, 2 and 4.

**U-Net** U-Net [19] is an excellent FCNN structure for semantic segmentation. This structure was proposed for biomedical image segmentation. It makes full use of encoders' information with concatenation structure thus achieves finer boundary quality and higher accuracy than the skip connection structure. The concatenation structure is demonstrated in Figure 5.

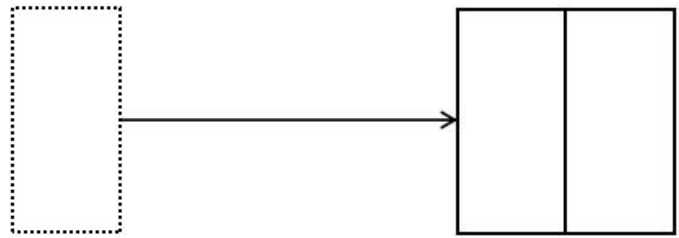

**Figure 5:** Concatenation structure U-Net uses in its decoder. The width of blocks in the figure represents channels in a single feature map. As the figure demonstrated, channels from every encoder's feature map are concatenated to channels from their counterpart decoder's feature map.

**ResNet50 + skip connection structure** ResNet [23] was proposed to solve gradient vanishing and feature map vanishing problem when training excessive deep CNN. The mechanism behind the ResNet is that identity connection [24] between non-adjacent layers does not influence the ideal mapping we want to achieve. However, the identity connection also make the back propagation more fluent because of additional short cut path the gradients can pass through. We find some of the mathematical deductions proving the residual structures' effectiveness. Therefore, we try to replace the VGG16 structure with ResNet50 structure. Considering the memory and time limitation, we did not try ResNet101 and ResNet152 as we do not have a huge data set and the paper presented the improvement is limited as the networks were deepened. The specific structure is presented in Figure 6.

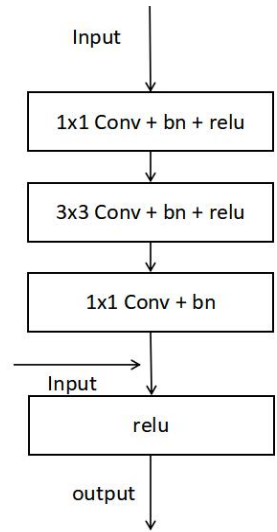

**Figure 6:** The residual unit ResNet50 uses. Input dose not only passes through 3 convolutional and batch normalization layers, it is also added to the output through an identity connection.

**Residual U-Net** The next step we take is applying residual structure to U-Net. We replace U-Net's convolutional layers with residual stacks each of which consist of 2 convolutional layers both in encoder and decoder. Besides the residual structure, we also insert batch normalization [25] layers between every convolutional layer and activation layer as it is reported the whiten data distribution accelerates the training process. The residual unit we used is presented in Figure 7.

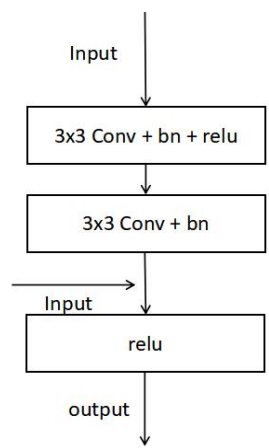

**Figure 7:** The residual unit we use in Residual U-Net. This type of structure replaces every 2 convolutional layers in U-Net.

## 3.2 Hierarchical Dice Loss

We now discuss the classifier we use to every feature extractor mentioned above. At first, we decide to use a regular cross-entropy loss using softmax function according to every pixel were originally labeled to make a 5-class classifier. The function can be defined as:

$$L_i = -\log(\frac{e^{f_{yi}}}{\sum_j e^{f_j}})$$

However, as the ratio of 5 class are nearly 2262:2:16:7:1, most of the pixels have a low loss as the first class is in dominant quantity and easy to train.

**Weighted Cross-entropy Loss** The simplest solution is giving a weight to every class. So the formula will become:

$$L_i = -w_{yi} \times \log(\frac{e^{f_{yi}}}{\sum_j e^{f_j}})$$

where the sum of weights equals 1. This loss function can balance the data set but it still remains a puzzle what the optimal distribution of weights is. In practice we give a list of 0.1, 0.35, 0.1, 0.1 and 0.35. We try the inverse numbers of their frequencies but it proves to be a over balance method.

**Bootstrapping loss** The idea that not every pixel is equal is clear in semantic segmentation. Works have been done showing that it is necessary to discriminate different pixels. Bootstrapping loss [26] is a good discrimination method because it only drops the worst pixels to train. Its formula is presented below:

$$L = -\frac{1}{\sum_i^N \sum_j^K 1\{y_i = j \text{ and } p_{ij} < t\}} \times (\sum_i^N \sum_j^K 1\{y_i = j \text{ and } p_{ij} < t\} \log p_{ij})$$

where t is a threshold, 1{ · } equals one when the condition inside the brackets holds, and otherwise equals zero. In practice we choose t = 0.9 so that it filters out 99.9% class 0 pixels which always give a convincing prediction.

**Sensitivity-Specificity Loss** Sensitivity and specificity are 2 main metrics when evaluate the segmentation results. Brosch et al. [27] adopted them into a loss function by minimizing the opposite number of weighted sensitivity and specificity. They referred the function as follows:

$$SS = \lambda \frac{\sum_{n=1}^N (r_n - p_n)^2 r_n}{\sum_{n=1}^N r_n + \varepsilon} + (1-\lambda) \frac{\sum_{n=1}^N (r_n - p_n)^2 (1-r_n)}{\sum_{n=1}^N (1-r_n) + \varepsilon}$$

where r is the label whether 1 or 0, p is the probability. $\lambda$ is weight to balance these 2 components, $\varepsilon$ is a small number to prevent denominators becoming zero.

**Dice Loss** Similarly when we try to use dice, namely dice score coefficient, to replace sensitivity and specificity, the model tends to perform better according to Milletari et al [28]. The function is as following:

$$DL = 1 - \frac{\sum_{n=1}^N p_n r_n + \varepsilon}{\sum_{n=1}^N p_n + r_n + \varepsilon} - \frac{\sum_{n=1}^N (1-p_n)(1-r_n) + \varepsilon}{\sum_{n=1}^N 2 - p_n - r_n + \varepsilon}$$

**Hierarchical Dice Loss** Both dice loss and sensitivity-specificity loss are powerful classification methods which can in a way balance the data sets. We suppose that dice loss's balancing ability derives from its denominators which includes sample numbers of every class. Therefore we try to apply dice loss to balance our data sets. However, the dice loss is designed for binary classification. Thus we decide to transfer the multiple classification into multiple binary classification. It is not easy to employ a multiple binary classification when the classes are independent with each other. However, the target classification task evolves classes with containing relationships. So we retain the 5-channel output layer of our FCNN, throw it to a softmax activation layer, add up corresponding probability of 5 labeled classes then get 3 target probabilities (p0, p1 and p2) which represent how much probable the pixel is compete tumor (sum of label 1, 2, 3, 4), tumor core (sum of label 1, 3, 4) and enhancing tumor (label 4). So the corresponding dice losses can be defined as follows:

$$DL_0 = 1 - \frac{\sum_{n=1}^{N} p_{n0} r_{n0} + \varepsilon}{\sum_{n=1}^{N} p_{n0} + r_{n0} + \varepsilon} - \frac{\sum_{n=1}^{N} (1 - p_{n0})(1 - r_{n0}) + \varepsilon}{\sum_{n=1}^{N} 2 - p_{n0} - r_{n0} + \varepsilon}$$

$$DL_1 = 1 - \frac{\sum_{n=1}^{N} p_{n1} r_{n1} + \varepsilon}{\sum_{n=1}^{N} p_{n1} + r_{n1} + \varepsilon} - \frac{\sum_{n=1}^{N} (p_{n0} - p_{n1})(r_{n0} - r_{n1}) + \varepsilon}{\sum_{n=1}^{N} p_{n0} + r_{n0} - p_{n1} - r_{n1} + \varepsilon}$$

$$DL_2 = 1 - \frac{\sum_{n=1}^{N} p_{n2} r_{n2} + \varepsilon}{\sum_{n=1}^{N} p_{n2} + r_{n2} + \varepsilon} - \frac{\sum_{n=1}^{N} (p_{n1} - p_{n2})(r_{n1} - r_{n2}) + \varepsilon}{\sum_{n=1}^{N} p_{n1} + r_{n1} - p_{n2} - r_{n2} + \varepsilon}$$

The final hierarchical dice loss is defined as the average of 3 separate dice losses.

$$DL_H = (DL_0 + DL_1 + DL_2) / 3$$

When making the classification, the classifier can be defined by four formulas. According to the 3 probabilities of one certain pixel, the necessary and sufficient conditions of non-tumor, complete tumor, tumor core and enhancing tumor are listed below:

$$p_{n0} < (1 - p_{n0})$$

$$p_{n0} > (1 - p_{n0})$$

$$p_{n1} > (p_{n0} - p_{n1}) \wedge p_{n0} > (1 - p_{n0})$$

$$p_{n2} > (p_{n1} - p_{n2}) \wedge p_{n1} > (p_{n0} - p_{n1}) \wedge p_{n0} > (1 - p_{n0})$$

# 4 Experiments

## 4.1 Train

**Down Sampling** Considering most of the pixels belong to class 0 and do little to improve distinguishing ability, we decide to down sample class 0. The method is simple: picking up only slices containing tumor pixels. Thus the ratio of non-tumor and tumor drop to 28:1 from 87:1.

**Deploying on Multiple GPU** We deploy same model on 3 GeForce GTX 1080 GPUs in Tensorflow. Each of the 3 GPUs works independently with their own input placeholders, forward stream and back propagation stream but shares same variables. Total loss are added up and the gradients are averaged by CPU.

**Implemention Details** We use Adaptive Moment Estimation (Adam) with a learning rate of 5e-5, learning rate decay of 0.95 per 10000 iterations, variables' moving average decay of 0.9999 and a batch size of 8 per GPU. We train different models using different structures and different loss functions and evaluate every of them after 100 epochs on LGG training data set.

## 4.2 Evaluation

In practice, we mainly use 4 metrics. They are precision, recall, mean intersection of union (mIoU) and dice. For convenience we draw tables for every FCNN architecture and different loss function then compare them with each other. Table 1 to Table 4 show the results from VGG + skip connection, U-Net, ResNet50 + skip connection and Residual U-Net with bootstrapping loss. Table 5 and 6 show the results from residual U-Net with original softmax cross-entropy loss and our hierarchical dice loss. All figures in the tables are the average values from LGG training data set.

We also give visual prediction results from 2 slices from 2 different 3D MRI images in Figure 8. It presents a contrast result between residual U-Net with softmax cross-entropy loss and residual U-Net with hierarchical dice loss after 100 epochs of training. It is clear that our model using hierarchical dice loss performs much better than its competitor. The model using cross-entropy loss can only approximately locate the complete tumor and part of tumor core. It fails to distinguish enhancing tumor in the 2 slices. On the contrary, model using hierarchical dice loss are prone to enhancing tumor, or in other words it gives more false positive predictions than we expected while classifying enhancing tumor. The model also fails to give boundaries between classes as fine as the ground truth images, which is a common consequence caused from mismatches between max-pooling layers and up sample layers. Despite of the shortcomings, it shows powerful and efficient distinguishing ability compared to the original design.

**Table 1:** Evaluation results from VGG + skip connection structure with bootstrapping loss. Comp Tumor: Complete Tumor (necrosis, edema, non-enhancing tumor and enhancing tumor), Tumor Core: Tumor Core (necrosis, non-enhancing tumor and enhancing tumor), Enh Tumor: Enhancing Tumor.

|            | Precision | Recall | mIoU   | Dice   |
|------------|-----------|--------|--------|--------|
| Comp Tumor | 0.8896    | 0.8714 | 0.7876 | 0.8778 |
| Tumor Core | 0.7550    | 0.8239 | 0.666  | 0.755  |
| Enh Tumor  | 0.1858    | 0.3361 | 0.1532 | 0.2203 |

**Table 2:** Evaluation results of U-Net with bootstrapping loss. Rows and columns are same to Table 1.

|            | Precision | Recall | mIoU   | Dice   |
|------------|-----------|--------|--------|--------|
| Comp Tumor | 0.9474    | 0.9281 | 0.8824 | 0.9366 |
| Tumor Core | 0.8422    | 0.9011 | 0.7768 | 0.8422 |
| Enh Tumor  | 0.4539    | 0.5684 | 0.3879 | 0.4804 |

**Table 3:** Evaluation results of ResNet50 + skip connection structure with bootstrapping loss. Rows and columns are same to Table 1.

|            | Precision | Recall | mIoU   | Dice   |
|------------|-----------|--------|--------|--------|
| Comp Tumor | 0.8907    | 0.9376 | 0.8409 | 0.9119 |
| Tumor Core | 0.7582    | 0.8643 | 0.6795 | 0.7582 |
| Enh Tumor  | 0.2153    | 0.4225 | 0.1861 | 0.2592 |

**Table 4:** Evaluation results of Residual U-Net with bootstrapping loss. Rows and columns are same to Table 1.

|            | Precision | Recall | mIoU   | Dice   |
|------------|-----------|--------|--------|--------|
| Comp Tumor | 0.9761    | 0.8691 | 0.8516 | 0.9152 |
| Tumor Core | 0.8602    | 0.9259 | 0.8073 | 0.8602 |
| Enh Tumor  | 0.4674    | 0.6423 | 0.4247 | 0.5112 |

**Table 5:** Evaluation results of Residual U-Net with original softmax loss. Rows and columns are same to Table 1.

|            | Precision | Recall  | mIoU   | Dice   |
|------------|-----------|---------|--------|--------|
| Comp Tumor | 0.7355    | 0.9373  | 0.6956 | 0.8067 |
| Tumor Core | 0.3740    | 0.7268  | 0.3414 | 0.3740 |
| Enh Tumor  | <1e-4     | <1e-4   | <1e-4  | <2e-4  |

**Table 6:** Evaluation results of Residual U-Net with hierarchical dice loss. Rows and columns are same to Table 1.

|            | Precision | Recall | mIoU   | Dice   |
|------------|-----------|--------|--------|--------|
| Comp Tumor | 0.9405    | 0.8767 | 0.8313 | 0.9050 |
| Tumor Core | 0.7871    | 0.8848 | 0.7119 | 0.8112 |
| Enh Tumor  | 0.4666    | 0.5645 | 0.4051 | 0.4938 |

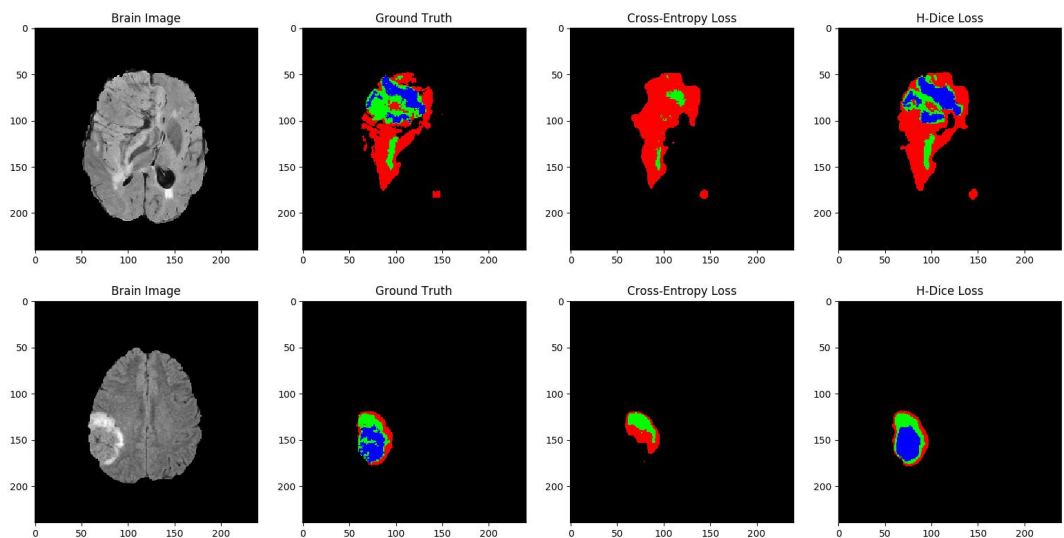

**Figure 8:** Visual results from 2 slices. Blue region represents enhancing tumor. Green region and blue region represent tumor core. The whole colored region represents complete tumor.

## 5 Discussion and Conclusion

In this paper, we designed refined fully convolutional neural networks with a hierarchical dice loss to segment subregions of brain tumor. Our adjustments include 2 parts. One of the adjustment is changing architectures of classic FCNN. We replace original convolutional layers in FCNN and U-Net with residual structures as well as batch normalization layers. Experiments prove that the residual FCNN gains higher value in all metrics than FCNN based on VGG16 net. Residual U-Net also performs better in most of the metrics especially when detecting enhancing tumor. This experiments' results indicate that residual structure or in other words, deeper convolutional nets have a powerful ability to extract features. Another comparison we made was U-Net and residual U-Net perform better than their counterparts using skip connection decoder. One of the main topic of semantic segmentation is how to combine local and global information. Through the experiments, the concatenate structure which U-Net uses seems to be a better choice. However, it dose not mean that the skip connection structure is inferior because the direct add-up method do not introduce extra dimensions thus decreases the memory usage and is easy to train. We tried several adjustments to the skip connection structure such as making a gradual up sample path which includes 4 2x up sample layers instead of 1 16x up sample layer, adding up all the feature maps produced by every max-pooling layer and retaining the high channel numbers of every encoder layers before the up sample layer. However, some of the adjustments make the model become too huge and some of the adjustments seem to impede converging. I believe further researches should be done to explore the potential of skip connection structure.

In this paper, the main problem we want to solve is how to balance the data set. Despite of the traditional method of down sampling the class in high frequency, we paid our attention to the loss function and classifier. We tried the multiple classifier at first and applied a bootstrapping loss function to neglect pixels which contribute little to the final loss but lower the mean value of loss. During training, we found when choosing the threshold of 0.9, the pixels which were not filtered out keep an acceptable ratio nearing to 2:2:5:4:1. The experiment results which have been demonstrated in section 4.2 prove that it elevates every metric values compared to the original softmax classifier and cross-entropy loss. On the other hand, we noticed that dice loss and sensitivity-specificity loss also have an ability to balance data set because the denominators of each components contain their class numbers. The mechanism is similar to adding a weight of class frequency's inverse number to the cross-entropy loss which is referred in section 3.2 as weighted cross-entropy loss. As the dice loss performs better in binary classification, we decide to convert the multiple classification to multiple binary classification. Then we have to design 3 hierarchical dice losses related and incorporated with each other as it is defined in section 4.2. However, although the hierarchical dice loss we proposed outperform much than the original cross-entropy loss, it dose not perform better than bootstrapping loss when comparing Table 4 and Table 6. Works still have be to done to improve the hierarchical dice loss. We are considering to add weights to 3 hierarchical losses or add weights to manually balance the pair components in a single dice loss.

Because of time limitation, we did not test the model on testing data set. Although we believe the evaluation of training data set is sufficient to reflect different models' and classifiers' ability, we decide to test our models on provided test data set in the future research.

We are still considering to implement a 3D FCNN to segment the whole 3D MRI image in 4 channel. Researchers have attempted to achieve a 3D segmentation or some of the researchers analyse 2D slices through 3 axis to give a holistic result [12]. We believe 3D segmentation should be more effective and more accurate because the receptive fields of elements in 3D FCNN should be much larger than 2D FCNN. We decide to achieve 3D semantic segmentation aiming to solve the problem of huge memory demand and make a trade-off between accuracy and recourse consuming.

# Reference

[1] Bauer, S., Wiest, R., Nolte, L. P., & Reyes, M. (2013). A survey of mri-based medical image analysis for brain tumor studies. *Physics in Medicine & Biology, 58*(13), R97.

[2] Havaei, M., Davy, A., Warde-Farley, D., Biard, A., Courville, A., & Bengio, Y., et al. (2017). Brain tumor segmentation with deep neural networks. *Medical Image Analysis, 35*, 18-31.

[3] Ali, N., Direko, Lu, C., & Ah, M. (2016). *Review of MRI-based Brain Tumor Image Segmentation Using Deep Learning Methods*. Elsevier Science Publishers B. V.

[4] Gordillo, N., Montseny, E., & Sobrevilla, P. (2013). State of the art survey on mri brain tumor segmentation. *Magnetic Resonance Imaging, 31*(8), 1426-38.

[5] Saritha, M., Joseph, K. P., & Mathew, A. T. (2013). Classification of mri brain images using combined wavelet entropy based spider web plots and probabilistic neural network. *Pattern Recognition Letters, 34*(16), 2151-2156.

[6] Kamnitsas, K., Ledig, C., Newcombe, V. F. J., Simpson, J. P., Kane, A. D., & Menon, D. K., et al. (2017). Efficient multi-scale 3d cnn with fully connected crf for accurate brain lesion segmentation. *Medical Image Analysis, 36*, 61.

[7] Wang, G., Li, W., Ourselin, S., & Vercauteren, T. (2017). Automatic brain tumor segmentation using cascaded anisotropic convolutional neural networks.

[8] Zöllner FG, Emblem KE, & Schad LR. (2012). Svm-based glioma grading: optimization by feature reduction analysis. *Zeitschrift Fur Medizinische Physik, 22*(3), 205-214.

[9] Havaei, M., Larochelle, H., Poulin, P., & Jodoin, P. M. (2016). Within-brain classification for brain tumor segmentation. *International Journal of Computer Assisted Radiology & Surgery, 11*(5), 777-788.

[10] Kwon, D., Shinohara, R. T., Akbari, H., & Davatzikos, C. (2014). Combining generative models for multifocal glioma segmentation and registration. (Vol.17, pp.763). Med Image Comput Comput Assist Interv.

[11] Zikic, D., Glocker, B., Konukoglu, E., Criminisi, A., Demiralp, C., & Shotton, J., et al. (2012). Decision forests for tissue-specific segmentation of high-grade gliomas in multi-channel mr. *Med Image Comput Comput Assist Interv, 15*(Pt 3), 369-376.

[12] Urban, G., Bendszus, M., Hamprecht, F. A., & Kleesiek, J. (2014). Multi-modal brain tumor segmentation using deep convolutional neural networks. *MICCAI BraTS (Brain Tumor Segmentation) Challenge. Proceedings, winning contribution*.

[13] Zikic, D., Ioannou, Y., Brown, M., & Criminisi, A. (2014). Segmentation of Brain Tumor Tissues with Convolutional Neural Networks. *Miccai Workshop on Multimodal Brain Tumor Segmentation Challenge*.

[14] Pereira, S., Pinto, A., Alves, V., & Silva, C. A. (2016). Brain tumor segmentation using convolutional neural networks in mri images. *IEEE Transactions on Medical Imaging, 35*(5), 1240-1251.

[15] Krizhevsky, Alex, Sutskever, Ilya, & Hinton, Geoffrey E. (2012). Imagenet classification with deep convolutional neural networks. *Communications of the Acm, 60*(2), 2012.

[16] Simonyan, K., & Zisserman, A. (2014). Very deep convolutional networks for large-scale image recognition. *Computer Science*.

[17] Long, J., Shelhamer, E., & Darrell, T. (2015). Fully convolutional networks for semantic segmentation. *Computer Vision and Pattern Recognition*(Vol.79, pp.3431-3440). IEEE.

[18] Badrinarayanan, V., Kendall, A., & Cipolla, R. (2017). Segnet: a deep convolutional encoder-decoder architecture for scene segmentation. *IEEE Transactions on Pattern Analysis & Machine Intelligence, PP*(99), 1-1.

[19] Ronneberger, O., Fischer, P., & Brox, T. (2015). *U-Net: Convolutional Networks for Biomedical Image Segmentation. Medical Image Computing and Computer-Assisted Intervention — MICCAI 2015*. Springer International Publishing.

[20] Huang, G., Liu, Z., Laurens van der Maaten, & Weinberger, K. Q. (2016). Densely connected convolutional networks.

[21] Pohlen, T., Hermans, A., Mathias, M., & Leibe, B. (2017). Full-resolution residual networks for semantic segmentation in street scenes.

[22] Szegedy, C., Liu, W., Jia, Y., Sermanet, P., Reed, S., & Anguelov, D., et al. (2015). Going deeper with convolutions. *Computer Vision and Pattern Recognition* (pp.1-9). IEEE.

[23] He, K., Zhang, X., Ren, S., & Sun, J. (2015). Deep residual learning for image recognition. 770-778.

[24] He, K., Zhang, X., Ren, S., & Sun, J. (2016). Identity mappings in deep residual networks. 630-645.

[25] Ioffe, S., & Szegedy, C. (2015). Batch normalization: accelerating deep network training by reducing internal covariate shift. 448-456.

[26] Wu, Z., Shen, C., & Hengel, A. V. D. (2016). Bridging category-level and instance-level semantic image segmentation.

[27] Brosch, T., Yoo, Y., Tang, L. Y. W., Li, D. K. B., Traboulsee, A., & Tam, R. (2015). *Deep Convolutional Encoder Networks for Multiple Sclerosis Lesion Segmentation. Medical Image Computing and Computer-Assisted Intervention — MICCAI 2015*. Springer International Publishing.

[28] Milletari, F., Navab, N., & Ahmadi, S. A. (2016). V-Net: Fully Convolutional Neural Networks for Volumetric Medical Image Segmentation. *Fourth International Conference on 3d Vision* (pp.565-571). IEEE.